\documentclass[a4paper,12pt]{article}
\usepackage{amsthm,amsmath,amssymb,authblk,bbm,geometry,epsfig,
enumerate,comment,nicefrac,listings,latexsym,mathrsfs,mathtools,
rotating,float,subeqnarray,cases,indentfirst,graphicx,
subfigure,amsfonts,booktabs,bm,epstopdf,outlines}
\begin{document}
\title{The Deep Ritz method: A deep learning-based 
numerical algorithm for solving variational problems}

\author[1]{ Weinan E}
\author[2]{Bing Yu}

\affil[1]{The Beijing Institute of Big Data Research\\
Department of Mathematics and PACM, Princeton University\\
School of Mathematical Sciences and BICMR, Peking University}
\affil[2]{School of Mathematical Sciences, Peking University}

\maketitle

%\begin{abstract}
\noindent{\bf Abstract} We propose a deep learning based method, the Deep Ritz Method, for numerically solving variational problems, particularly the ones that arise from partial differential equations. The Deep Ritz method is naturally nonlinear, naturally adaptive and has the potential to work in rather high dimensions. The framework is quite simple and fits well with the stochastic gradient descent method used in deep learning. We illustrate the method on several problems including some eigenvalue problems.
%\end{abstract}

\section{Introduction}

Deep learning has had great success in computer vision and other artificial
intelligence tasks \cite{book}. Underlying this success is a new way to
approximate functions, from an additive construction commonly used in
approximation theory to a compositional construction used in 
deep neural networks.
The compositional construction seems to be particularly powerful
in high dimensions. This suggests that deep neural network based models
can be of use in other contexts that involve constructing functions.
This includes solving partial differential
equations, molecular modeling, model reduction,  etc.
These aspects have been explored recently in \cite{E, Han-Jentzen-E,
E-Han-Jentzen, Beck-E-Jentzen, Han-Zhang-Car-E, Zhang-Han-Wang-Car-E}. 

In this paper, we continue this line of work and propose a new
algorithm for solving variational problems. We call this new algorithm
the Deep Ritz method since it is based on using the neural
network representation of functions in the context of the Ritz method.
The Deep Ritz method has a number of interesting and promising features,
which we explore later in the paper.

\section{The Deep Ritz Method}
%\label{sec:method}

An explicit example of the kind of variational problems we are interested
in is \cite{Evans}
\begin{equation}
\min_{u \in H} I(u)
\end{equation}
where
\begin{equation}
I(u) = \int_{\Omega} \left( \frac 12  |\nabla u(x)|^2 -  f(x) u(x) \right) dx
\label{I(u)}
\end{equation}
and $H$ is the set of admissible functions (also called
trial function, here represented by $u$), 
$f$ is a given function, representing
external forcing to the system under consideration.
Problems of this type are fairly common in physical sciences.
The Deep Ritz method is based on the following set of ideas:

\begin{enumerate}
\item Deep neural network based approximation of the trial function.
\item A numerical quadrature rule for the functional.
\item An algorithm for solving the final optimization problem.
\end{enumerate}

\subsection{Building trial functions}

The basic component of the Deep Ritz method is a nonlinear transformation $ x \rightarrow z_\theta (x)\in \mathbb{R}^{m} $ 
defined by a deep neural network.
Here $\theta$ denotes the parameters, typically the weights in the 
neural network, that help to define this transformation.
In the architecture that we use, each layer of the network is constructed by stacking several blocks, each block consists of two linear transformations, 
two activation functions and a residual connection, 
both the input $s$ and the output $t$ of the block are vectors in $\mathbb{R}^{m}$. 
The $i$-th block can  be expressed as:
\begin{equation}
t=f_i(s)=\phi(W_{i,2} \cdot \phi(W_{i,1} s +b_{i,1})+b_{i,2})+s
\label{block}
\end{equation}
where $W_{i,1},W_{i,2} \in \mathbb{R}^{m \times m}$, $b_{i,1},b_{i,2} \in \mathbb{R}^{m}$
are parameters associated with the block. 
$\phi$ is the (scalar) activation function \cite{book}.

\begin{figure}[!h]
\centering
\vspace{2pt}
%\subfigure[Solution of Deep Ritz method, $811$ parameters]{
\includegraphics[width=0.45\textwidth,height=0.6\textwidth]{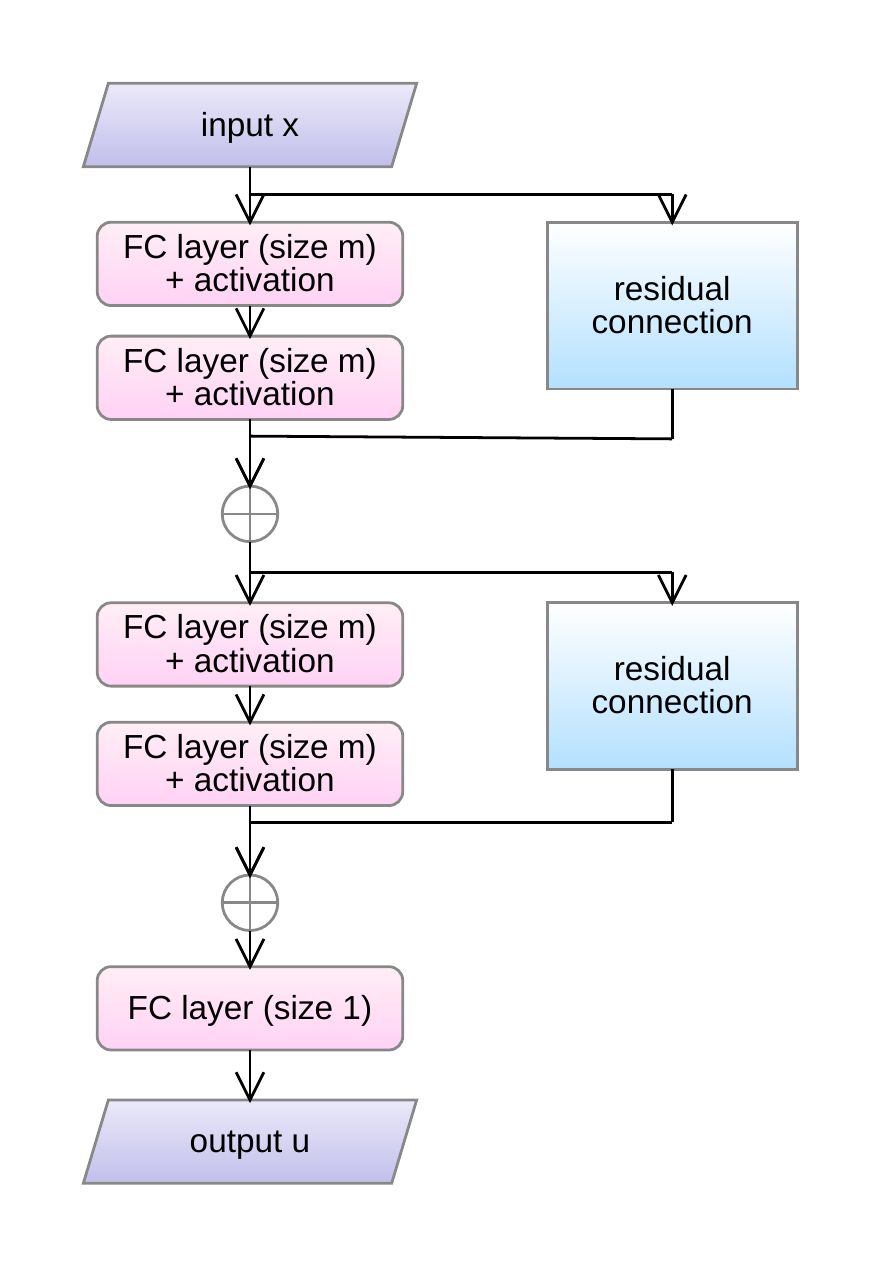}
%\label{fig:net_structure_subfig}
%}
\caption{The figure shows a network with two blocks and an output linear layer.
Each block consists of two fully-connected layers and a skip connection.\label{fig:net_structure_1}}
\end{figure}

Our experience has suggested that the smoothness of the activation
function $\phi$ plays a key role in the accuracy of the algorithm.
To balance simplicity and accuracy, we have decided to use
\begin{equation}
\phi(x)=\max\{x^3,0\}
\end{equation}
The last term in \eqref{block}, the residual connection, 
makes the network much easier to train since it helps to avoid the vanishing gradient problem \cite{resnet}. 
The structure of the two blocks, including two residual connections, 
is shown in Figure \ref{fig:net_structure_1}.
 
%{\bf can you explain more about the residual connection?}

%{\bf also, it is useful to draw a figure to illustrate the block}

%{\bf I have explained why res-connections are helpful, cited the resnet paper and added a network architecture figure.}

The full $n$-layer network can now be expressed as:
\begin{equation}
z_{\theta}(x)=f_n\circ ... \circ f_1(x)
\end{equation}
$\theta$ denotes the set of all the parameters in the whole network.
Note the input $x$ for the first block is in $\mathbb{R}^d$, not $\mathbb{R}^{m}$.
To handle this discrepancy we can either
pad $x$ by a zero vector when $d<m$, or apply a linear transformation 
on $x$ when $d>m$. 
Having $z_\theta$, we obtain $u$ by
\begin{equation}
u(x; \theta) = a \cdot z_\theta (x) + b
\label{function-form}
\end{equation}
Here in the left-hand side and in what follows, we will use
$\theta$ to denote the full parameter set $\{\theta, a, b \}$. Substituting this into the form of $I$, we obtain a function of $\theta$,
which we should minimize.

For the functional that occurs in \eqref{I(u)}, denote:
\begin{equation}
g(x; \theta) = \frac 12 |\nabla_x u(x; \theta)|^2 - f(x) u(x; \theta)
\end{equation}
then we are left with the optimization problem:
\begin{equation}
\min_\theta L(\theta), \quad L(\theta) = \int_\Omega g(x; \theta) dx
\label{integral}
\end{equation}

\subsection{The stochastic gradient descent algorithm and the quadrature rule}

To finish describing the algorithm, we need to furnish the remaining
two components:  the optimization algorithm and the discretization
of the integral in $I$ in \eqref{I(u)} or $L$ in \eqref{integral}. 
The latter is necessary since computing the integral
in $I$ (or $L$) explicitly for functions of the form 
\eqref{function-form} is quite an impossible task.

In machine learning, the optimization problem that one encounters
often takes the form:
\begin{equation}
\min_{x\in \mathbb{R}^d}\quad  L(\theta) := \frac{1}{N}\sum_{i=1}^N L_i(\theta),
\end{equation}
where each term at the right-hand side corresponds to one data point.
$n$, the number of data points, is typically very large.
For this problem, the algorithm of choice is the 
stochastic gradient descent (SGD) method, which can be 
described as follows:
\begin{equation}
\theta^{k+1} = \theta^{k} - \eta \nabla f_{\gamma^k}(\theta^k). 
\label{eq:sga_iter}
\end{equation}
Here $\{\gamma^k\}$ are i.i.d random variables uniformly distributed over $\{1,2,\cdots, n\}$.
This is the stochastic version of the gradient descent algorithm (GD).
The key idea is that instead of computing the sum when evaluating the
gradient of $L$, we simply randomly choose one term in the sum.
Compared with GD, SGD requires only one function evaluation of $n$
function evaluations at each iteration.
In practice, instead of picking one term, one chooses a "mini-batch"
of terms at each step. 

At a first sight, our problem seems different from the ones that
occur in machine learning since there are no data involved. 
The connection becomes clear once we view the integral in $I$ as
a continuous sum, each point in $\Omega$ then becomes a data point.
Therefore, at each step of the SGD iteration, one chooses
a mini-batch of points to discretize the integral.
These points are chosen randomly and the same quadrature weight is 
used at every point.

Note that if we use standard quadrature rules to discretize the integral,
then we are bound to choose a fixed set of nodes.
In this case, we run into the risk where
the integrand is minimized on these fixed nodes but the functional
itself is far from being minimized.  
It is nice that SGD fits naturally with the 
needed numerical integration in this context.

In summary, the SGD in this context is given by:
\begin{equation}
\theta^{k+1} = \theta^{k} - \eta \nabla_\theta \frac 1 N
\sum_{j=1}^{N} g(x_{j, k}; \theta^k)
\label{eq:sga_iter_var}
\end{equation}
where for each $k$, $\{x_{j, k} \}$ is a set of points in $\Omega$
that are randomly sampled with uniform distribution. To accelerate 
the training of the neural network, we use the Adam optimizer version of the
SGD \cite{adam}.

%{\bf do we need to explain the Adam optimizer?}

%{\bf I cite the paper about Adam optimizer here}

\section{Numerical Results}

\subsection{The Poisson equation in two dimension} 

Consider the Poisson equation:
\begin{equation}
\begin{aligned}
-&\Delta u(x)=1,\quad &x\in \Omega\\
&u(x)=0,\quad &x\in \partial \Omega
\end{aligned}
\end{equation}
where $\Omega=(-1,1)\times (-1,1) \backslash [0,1)\times \{0\}$.
The solution to this problem suffers from the well-known "corner singularity"
caused by the nature of the domain \cite{Strang-Fix}.  
A simple asymptotic analysis shows that at the origin,  the solution
behaves as
$u(x)=u(r,\theta) \sim r^{\frac{1}{2}}\sin\frac{\theta}{2}$
\cite{Strang-Fix}.
Models of this type have been extensively used
to help developing and testing adaptive finite element methods.

The network we used to solve this problem is a stack of
four blocks (eight fully-connected layers) and an output layer
with $m=10$.  There are a total of $811$ parameters in the model.
As far as we can tell, this network structure is not special in any way. It is simply the one that we used.
%{\bf does this mean four layers?}

%{\bf I use the word ``block" here to stand for two layers and a skip connection.}

\begin{figure}[!h]
\centering
\vspace{2pt}
\subfigure[Solution of Deep Ritz method, $811$ parameters]{
\includegraphics[width=0.45\textwidth,height=0.3\textwidth]{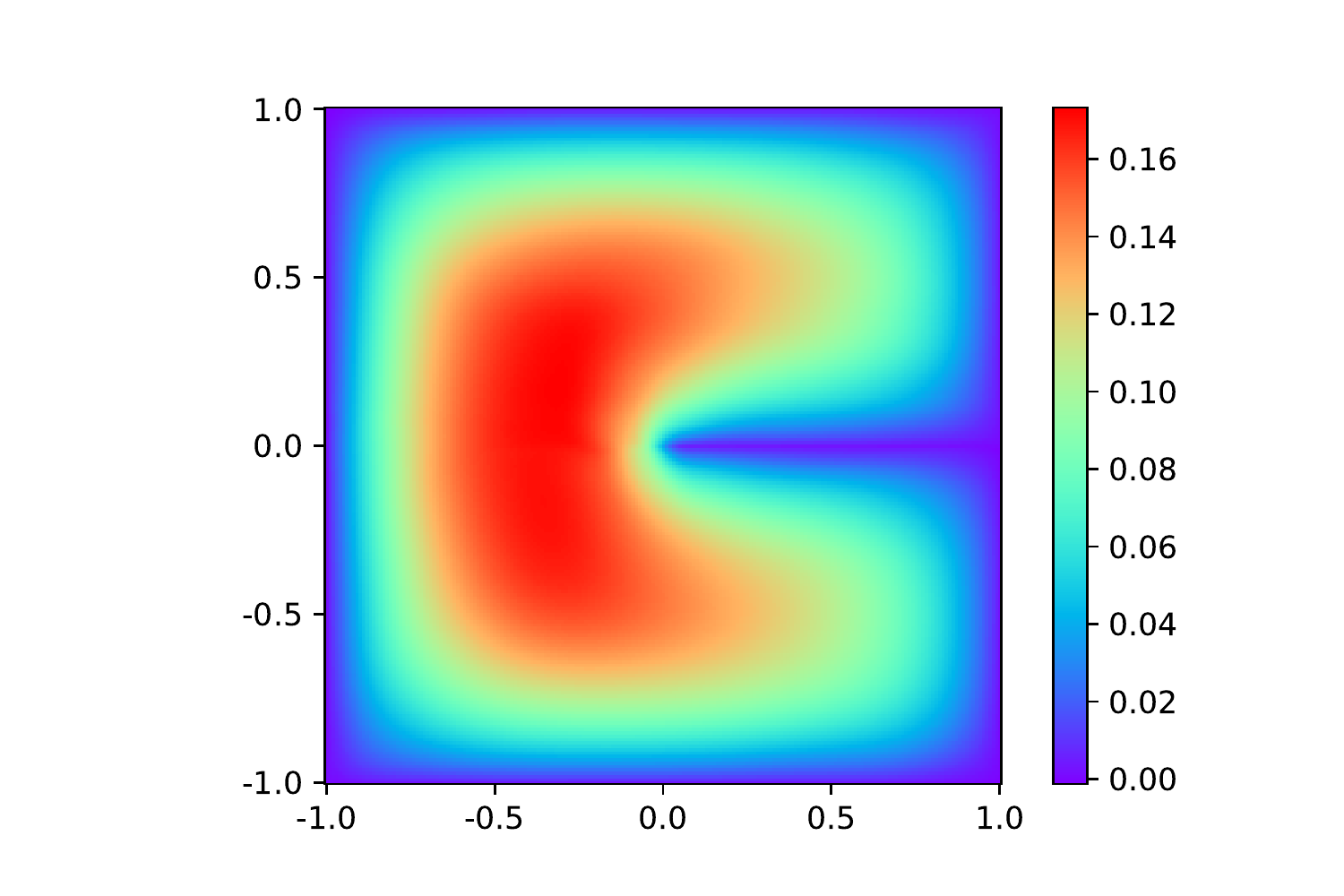}
\label{fig:fig_01}
}
\subfigure[Solution of finite difference method, $1,681$ parameters]{
\includegraphics[width=0.45\textwidth,height=0.3\textwidth]{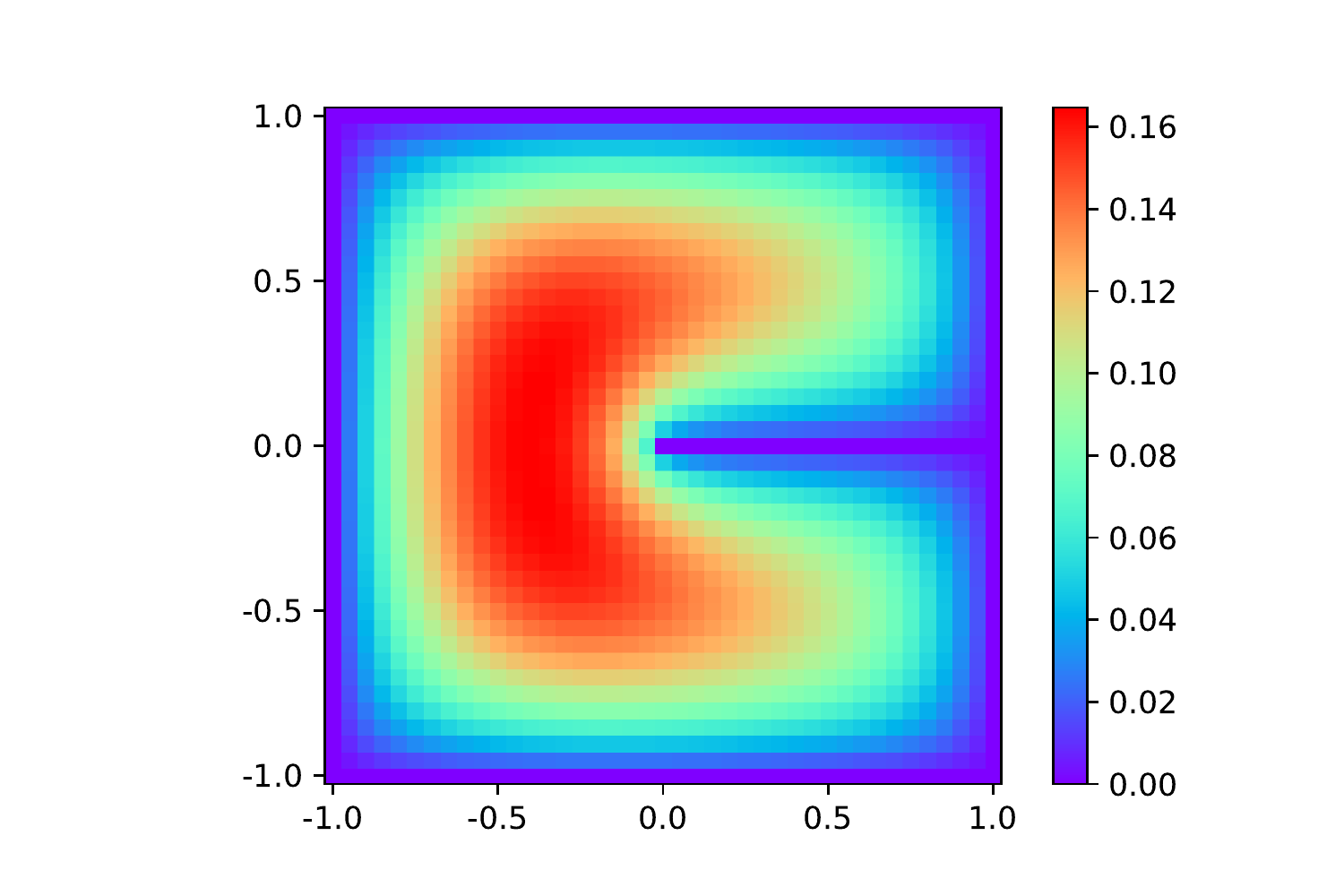}
\label{fig:fig_01_d}
}
\vspace{3pt}
\label{fig:01}
\caption{Solutions computed by two different methods.}
\end{figure}

The boundary condition causes some problems.
Here for simplicity, we use a penalty method and consider 
the modified functional
\begin{equation}
I(u) = \int_\Omega \left( \frac{1}{2} |\nabla_x u(x)|^2 - f(x) u(x) \right) dx
+\beta \int_{\partial \Omega} u(x)^2 ds 
\end{equation}
We choose $\beta = 500$.
The results from the Deep Ritz method is shown in
see Figure \ref{fig:fig_01}. For comparison, we also plot the result
of the finite difference method with $\Delta x_1=\Delta x_2=0.1$ ($1,681$ 
degrees of freedom), see Figure \ref{fig:fig_01_d}.

%{\bf maybe it is better to do contour plot,  it is important to exhibit the singularity of the solution}

%{\bf I have replaced the original surface figures with contour figures.}

To analyze the error more quantitatively, we consider the following
problem
\begin{equation}
\begin{aligned}
&\Delta u(x)=0,\quad &x\in \Omega\\
&u(x)=u(r,\theta)=r^{\frac{1}{2}}\sin\frac{\theta}{2},\quad &x\in \partial \Omega
\end{aligned}
\end{equation}
where $\Omega=(-1,1)\times (-1,1) \backslash [0,1)\times \{0\}$.
This problem has an explicit solution 
$u^*(x)=r^{\frac{1}{2}}\sin\frac{\theta}{2}$ in polar coordinates.
The error $e=\max |u^*(x)-u_h(x)|$, where $u^*$ and $u_h$ are the 
exact and approximate solutions respectively, is shown in Table \ref{tab:eq} for both the Deep Ritz method and the finite difference method (on uniform grids).
We can see that with fewer parameters, the Deep Ritz method 
gives more accurate solution than the finite difference method.

\begin{table}
\caption{Error of Deep Ritz method (DRM) and finite difference
method (FDM)}\label{tab:eq}
\begin{center}
\begin{tabular}{c|ccccc}
\hline
Method  & Blocks Num & Parameters & relative $L_2$ error \\ \hline
DRM & 3 & 591 & 0.0079\\ \hline
& 4 & 811 & 0.0072\\ \hline
& 5 & 1031 & 0.00647\\ \hline
& 6 & 1251 & 0.0057\\ \hline
FDM & & 625 & 0.0125\\ \hline
& & 2401 & 0.0063\\ \hline
\end{tabular}
\end{center}
\end{table}

%\begin{figure}[!h]
%\centering
%\vspace{2pt}
%\includegraphics[width=0.45\textwidth,height=0.4\textwidth]{fig/lnerr.png}
%\vspace{3pt}
%\label{fig:err_example1}
%\caption{The two figures shows the error and ln(error) during the training process, the x-axis represents for iteration steps (one stands for 100 steps). The $L_2$ error is computed on $25 \cdot 25$ grid points every 100 iterations.}
%\end{figure}

%present the error as a function of the number of layers in the
%neural network.

%also, present the training time as the right hand side $f$ changes

Being a naturally nonlinear variational method, the Deep Ritz method
is also naturally adaptive. We believe that this contributes to the
better accuracy of the Deep Ritz method.

\subsection{Poisson equation in high dimension}

Experiences in computer vision and other artificial intelligence
tasks suggest that deep learning-based methods are particularly powerful
in high dimensions. This has been confirmed by the results of
the Deep BSDE method \cite{Han-Jentzen-E}.  
In this subsection, we investigate the performance
of the Deep Ritz method in relatively high dimension.

Consider  ($d=10$)
\begin{equation}
\begin{aligned}
&-\Delta u=0, \quad &x \in (0,1)^{10}\\
&u(x)=\sum_{k=1}^{5} x_{2k-1} x_{2k}, \quad &x \in \partial (0,1)^{10}\,.
\end{aligned}
\end{equation}
%{\bf please define the upper and lower limits of the sum in the boundary condition.}

%{\bf this problem is too simple, the solution is simply $ u = \sum x_{2k-1} x_{2k}$ }
The solution of this problem is simply $u(x)=\sum_{k=1}^{5} x_{2k-1} x_{2k}$, and we will use the exact solution to compute the error of our model later. 

%{\bf I have revised it.}

\begin{figure}[!h]
\centering
\vspace{2pt}
\subfigure[$\ln{e}$ and $\ln{loss_{boundary}}$, d=10]{
\includegraphics[width=0.45\textwidth,height=0.4\textwidth]{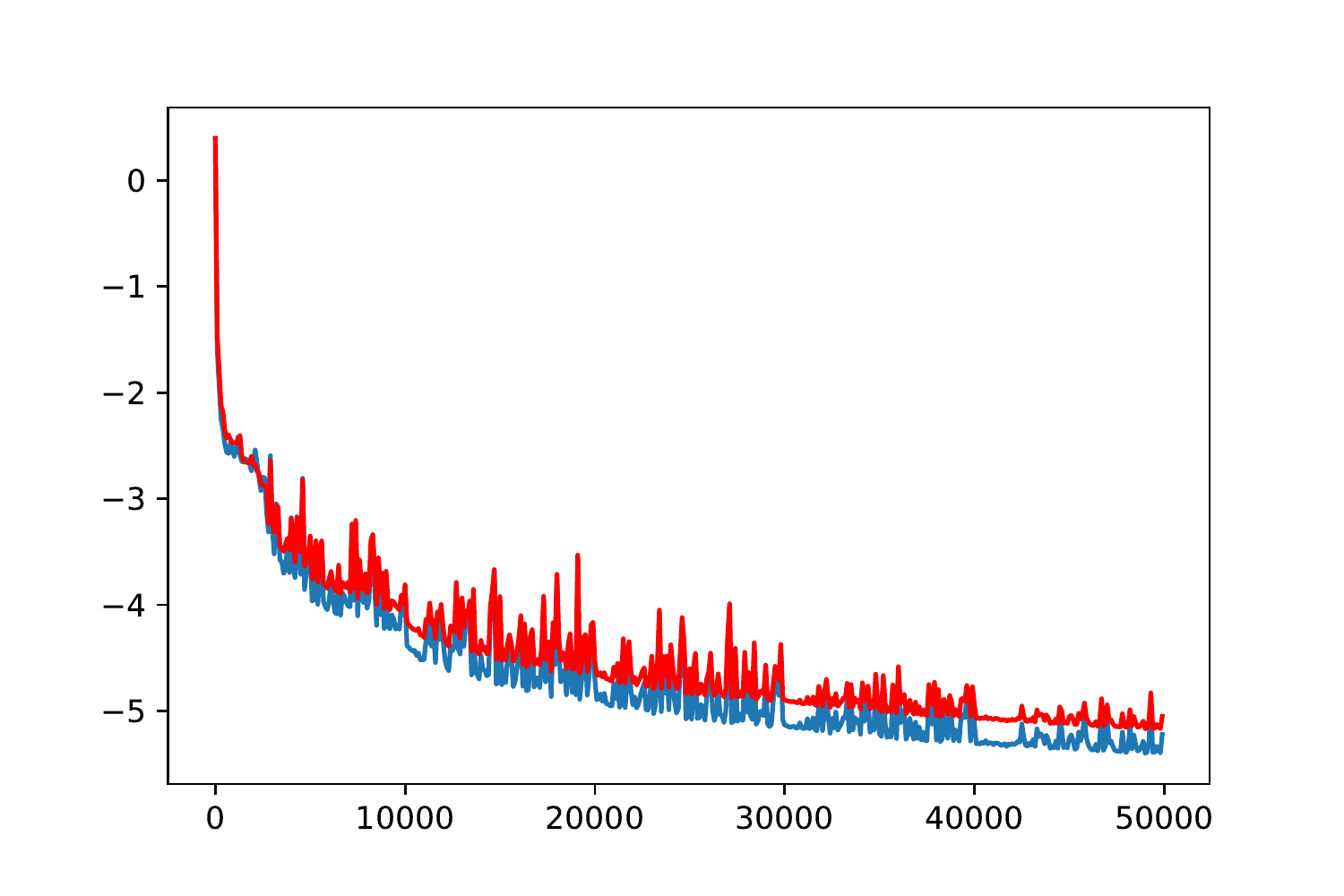}
\label{fig:err_10}
}
\subfigure[$\ln{e}$ and $\ln{loss_{boundary}}$, d=100]{
\includegraphics[width=0.45\textwidth,height=0.4\textwidth]{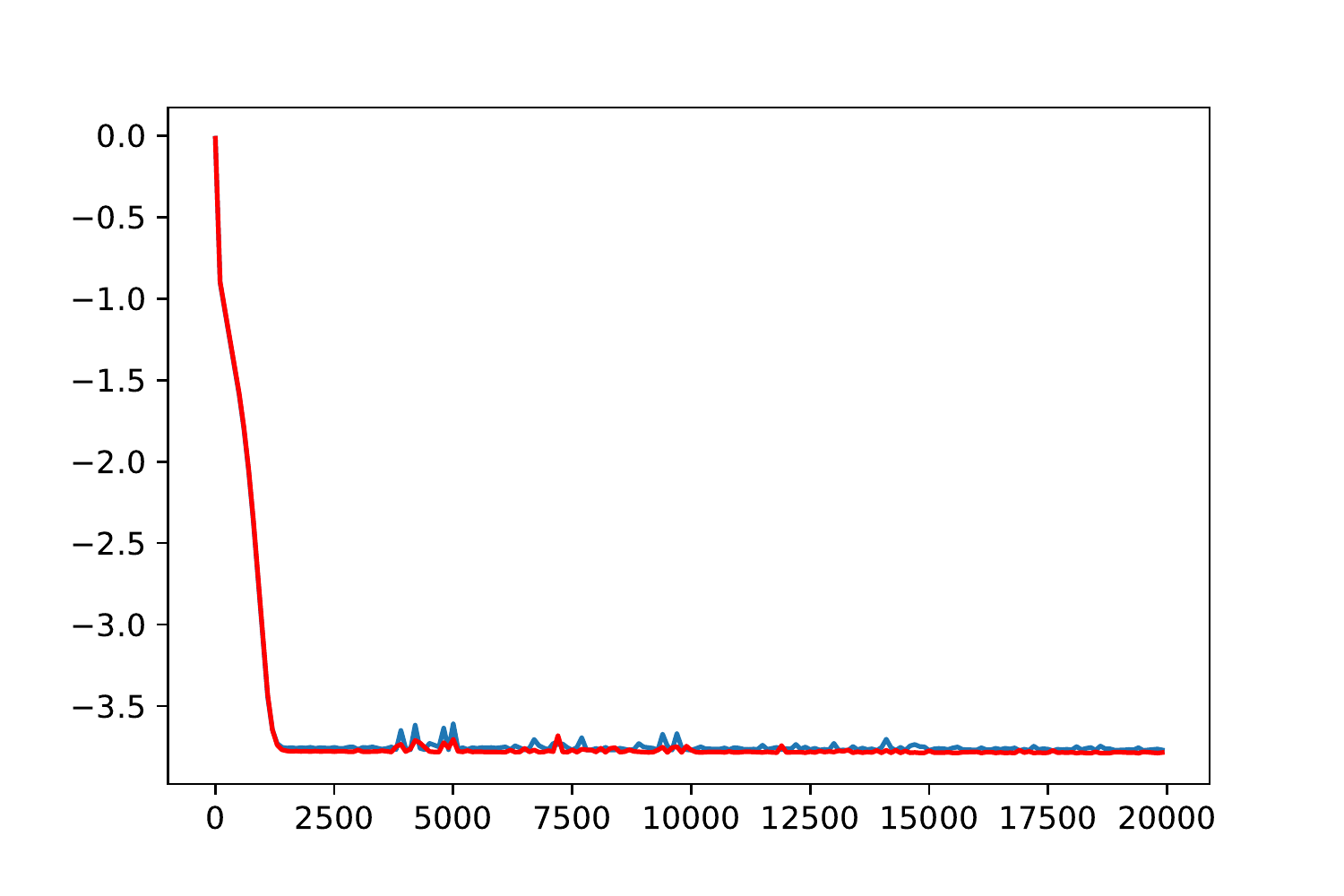}
\label{fig:err_100}
}
\vspace{3pt}

\caption{The total error and error at the boundary during the training 
process. The x-axis represents the iteration steps. 
The blue curves show the relative error of $u$.
The red curves show the relative error on the boundary. \label{fig:errk}
}
\end{figure}

For the network structure, we stack six fully-connected layers with three skip connections and a final linear layer, and there are a total of 671 parameters.
For numerical integration, at each step of the SGD iteration,
we sample 1,000 points in $\Omega$ and 100 points at each hyperplane that composes $\partial \Omega$. We set $\beta=10^3$. 
After 50,000 iterations, the relative $L_2$ error was reduced to
about $0.4\%$. The training process is shown in
Figure \ref{fig:err_10}.

%{\bf this needs to be made more clear}.

%{\bf I have explained it.}

Also shown in Figure  \ref{fig:err_100} is the training process for the
problem:
\begin{equation}
\begin{aligned}
-&\Delta u=-200 \qquad &x \in (0,1)^{d}\\
&u(x)=\sum_k x_k^2 \qquad &x \in \partial (0,1)^{d}
\end{aligned}
\end{equation}
with $d=100$ with a similar
network structure (stack 3 blocks of size m=100). The solution of this problem is $u(x)=\sum_k x_k^2$. After 50000 iterations, the relative error is reduced to about $2.2\%$.

\subsection{An example with the Neumann boundary condition}
Consider:
\begin{equation}
\begin{aligned}
-&\Delta u+\pi^2u=2\pi^2\sum_k \cos (\pi x_k) \qquad &x \in [0,1]^d\\
&\frac{\partial u}{\partial n}|_{\partial [0,1]^d}=0 &x \in \partial [0,1]^d
\end{aligned}
\end{equation}
The exact solution is $u(x)=\sum_k \cos(\pi x_k)$

\begin{figure}[!h]
\centering
\vspace{2pt}
\subfigure[$\ln{e}$, d=5]{
\includegraphics[width=0.45\textwidth,height=0.4\textwidth]{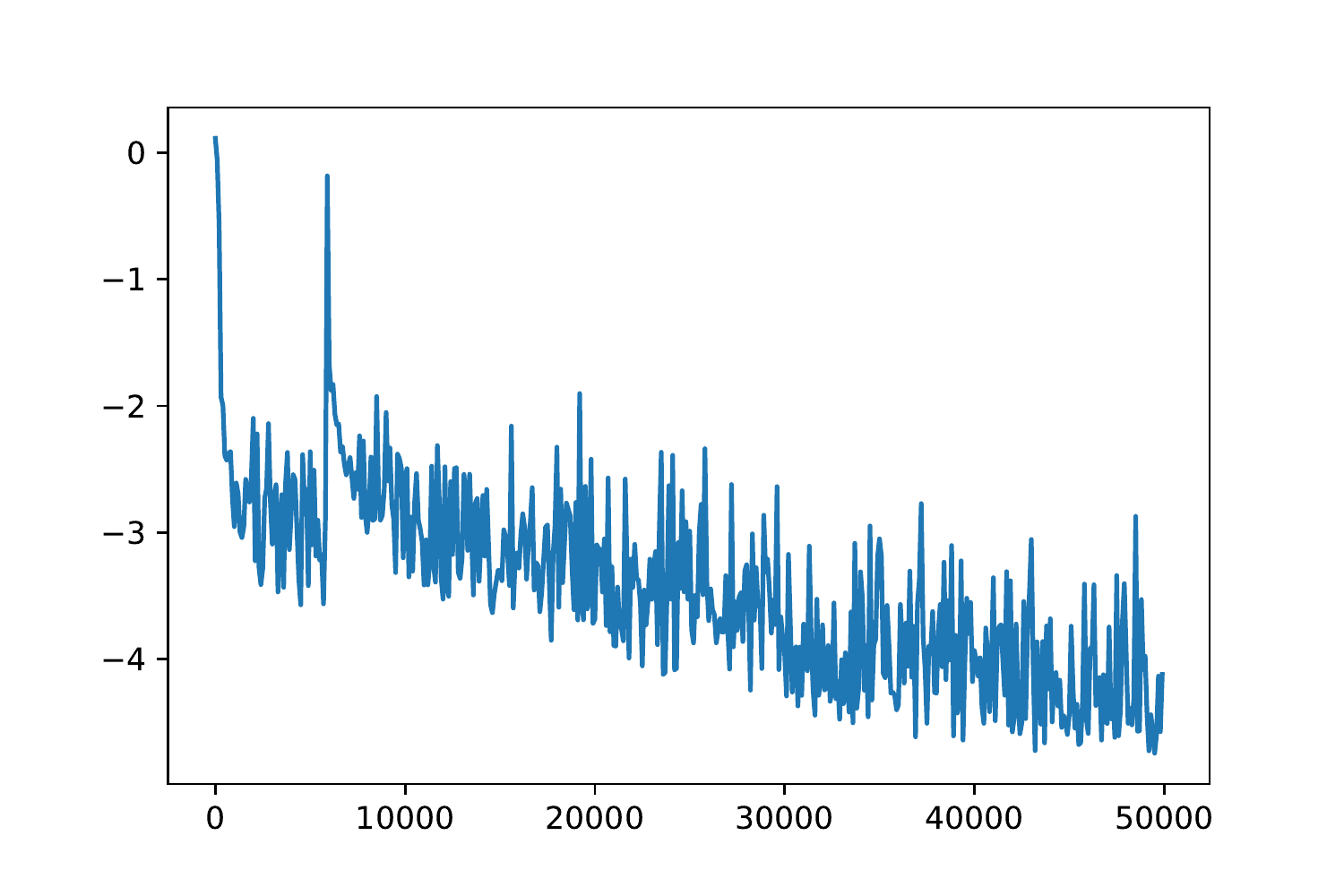}}
\subfigure[$\ln{e}$, d=10]{
\includegraphics[width=0.45\textwidth,height=0.4\textwidth]{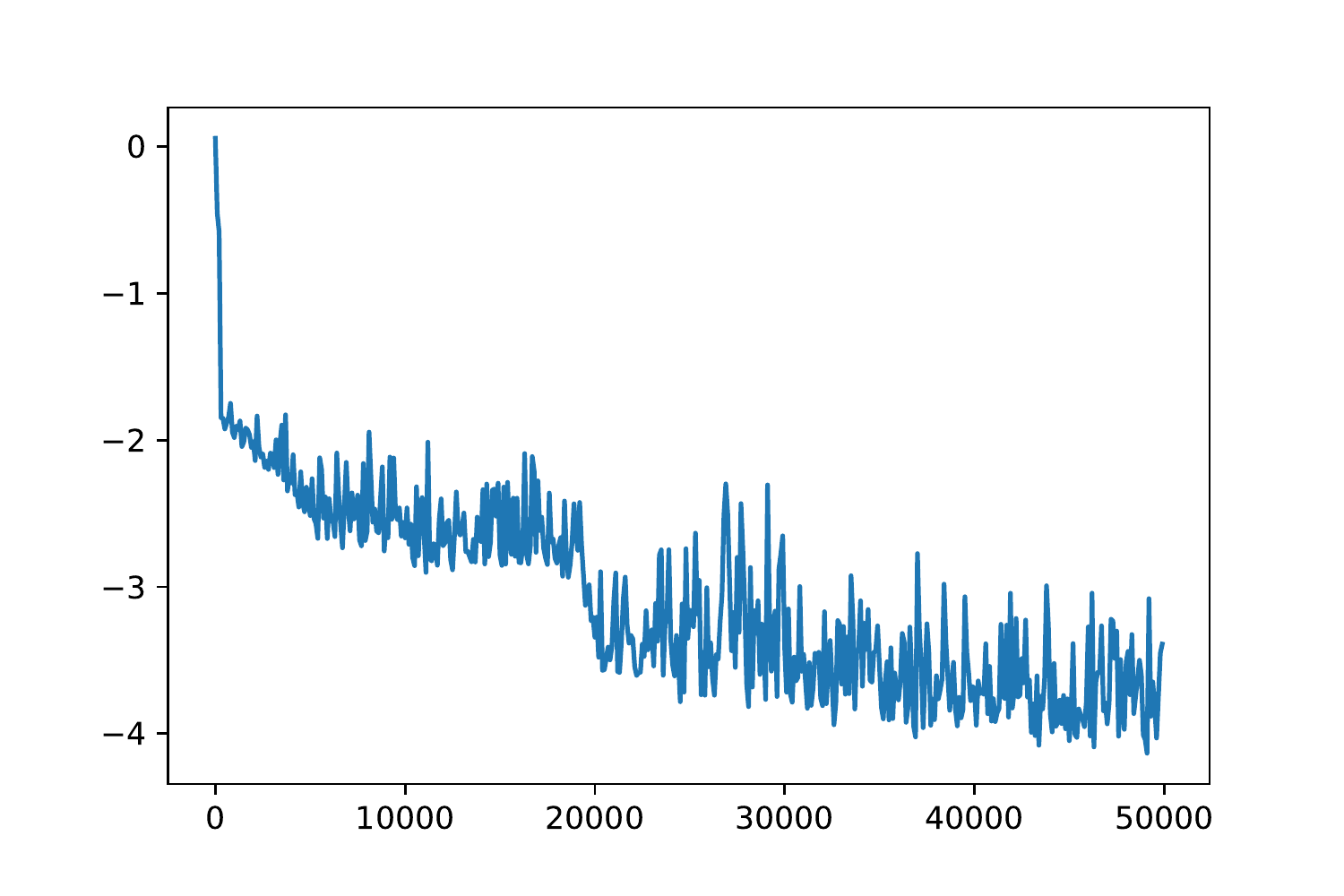}}
\vspace{3pt}
\caption{The error during the training process ($d=5$ and $d=10$).\label{fig:Neumann}
}
\end{figure}

In this case, we can simply use
$$
I(u) = \int_{\Omega} \left( \frac 12 \left( |\nabla u(x)|^2 +\pi^2 u(x)^2 \right)- 
 f(x) u(x) \right) dx
$$
without any penalty function for the boundary.

With a similar network structure the
relative $L_2$ error reaches $1.3\%$ for $d=5$ and $1.9\%$ for $d=10$. 
The training process is shown in Figure \ref{fig:Neumann}.

\subsection{Transfer learning}

An important component of the training process is the initialization.
Here we investigate the benefit of transferring weights in the network
when the forcing function $f$ is changed.

Consider the problem:
\begin{equation}
\begin{aligned}
-&\Delta u(x)=6(1+x_1)(1-x_1)x_2+2(1+x_2)(1-x_2)x_2 &x\in \Omega\\
&u(x)=r^{\frac{1}{2}}\sin{\frac{\theta}{2}}+(1+x_1)(1-x_1)(1+x_2)(1-x_2)x_2 \quad &x\in \partial \Omega
\end{aligned}
\end{equation}
%\begin{equation}
%u(x)=(x_1^2+x_2^2)^{\frac{1}{4}}(\frac{1}{2}(1-x_1{(x_1^2+x_2^2)^{-\frac{1}{2}}}))^{\frac{1}{2}}+(1+x_1)(1-x_1)(1+x_2)(1-x_2)x_2,\quad x\in \partial \Omega
%\end{equation}
where $\Omega=(-1,1)\times (-1,1) \backslash [0,1)\times \{0\}$.
Here we used a mixture of rectangular and polar coordinates.
The exact solution is $$u(x)=r^{\frac{1}{2}}\sin{\frac{\theta}{2}}+(1+x_1)(1-x_1)(1+x_2)(1-x_2)x_2$$.

\begin{figure}[!h]
\centering
\subfigure[$\ln {err}$]{
\includegraphics[width=0.45\textwidth,height=0.4\textwidth]{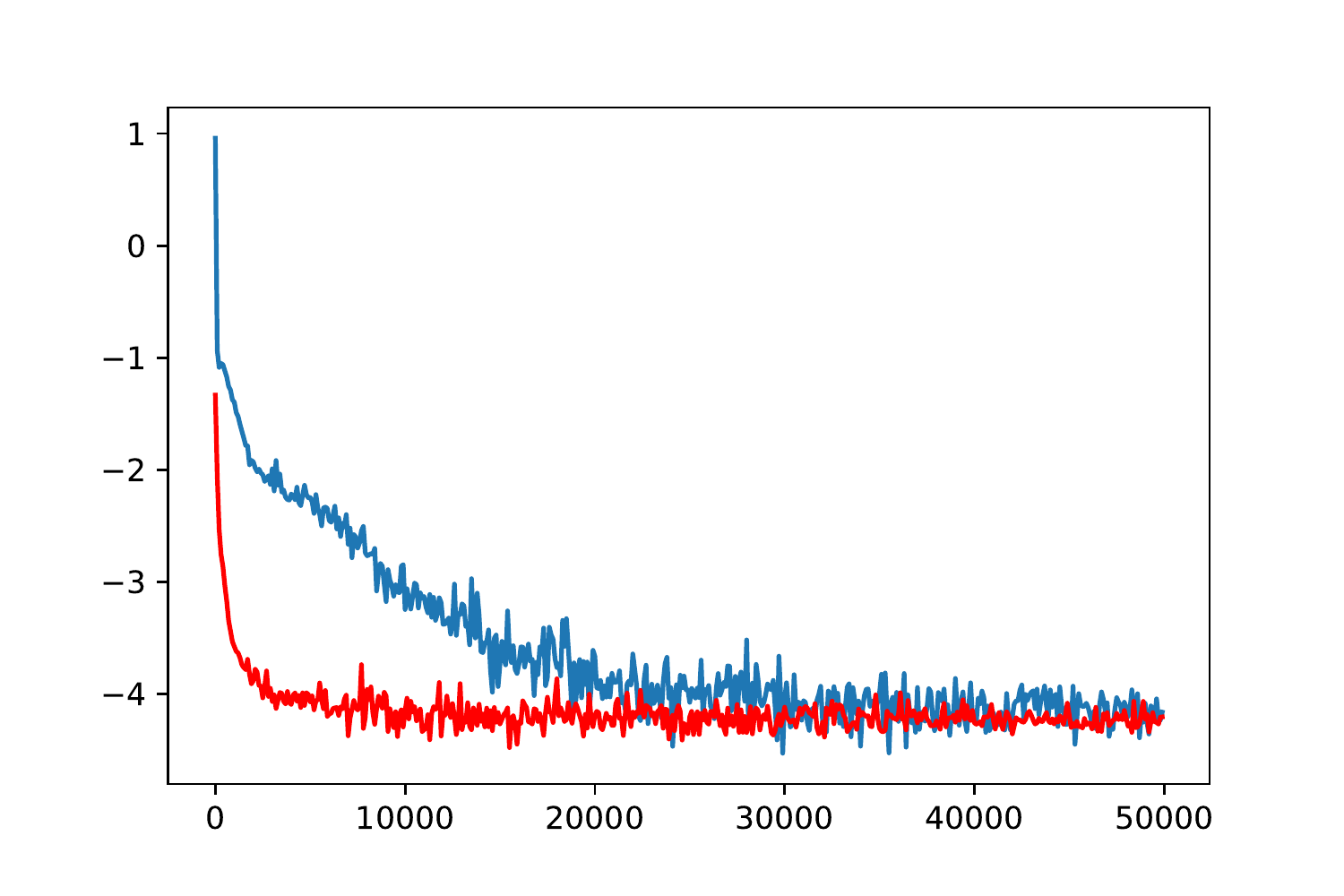}
\label{fig:transfer_a}
}
\subfigure[$\ln{||\Delta W||^2_2}$]{
\includegraphics[width=0.45\textwidth,height=0.4\textwidth]{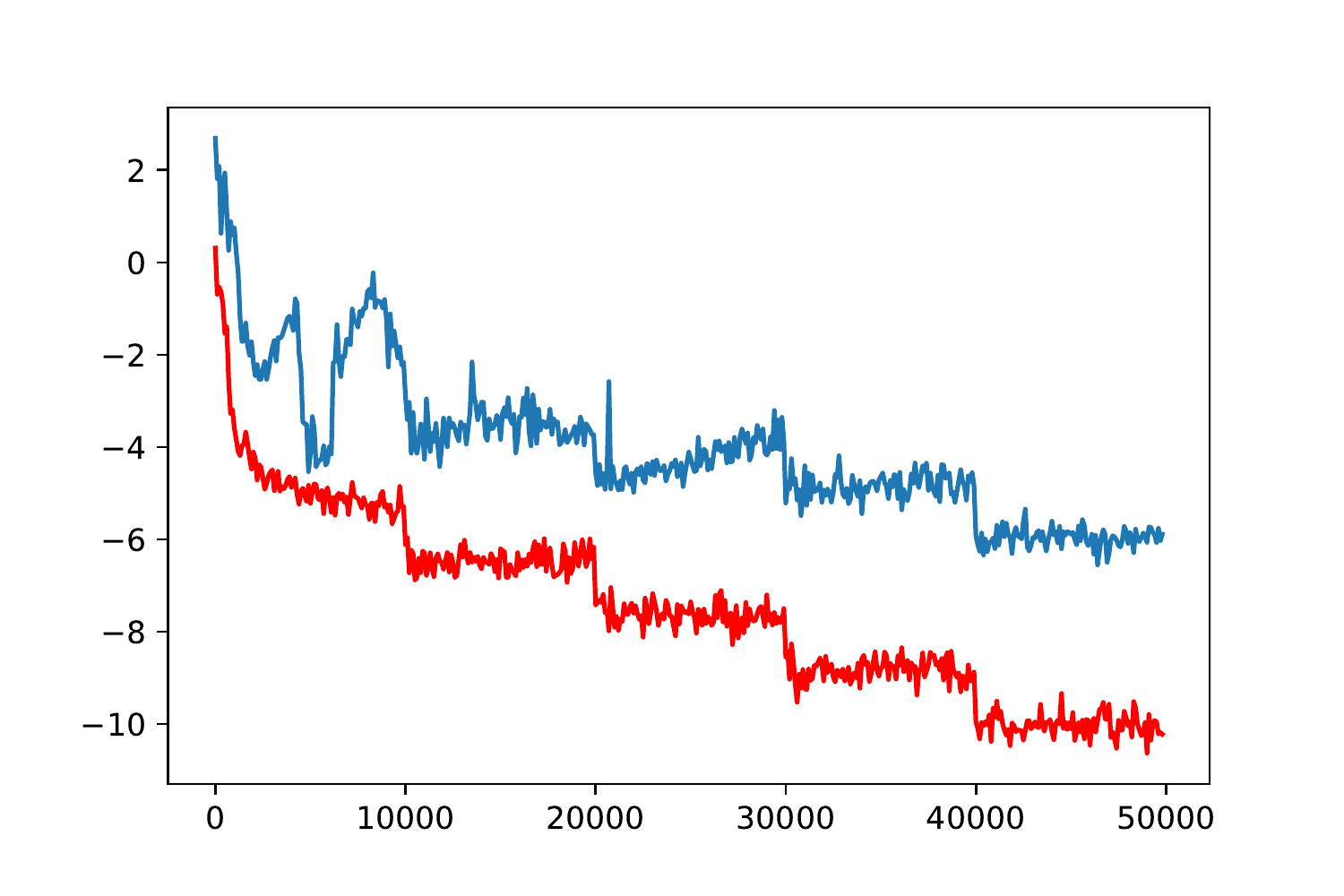}
\label{fig:transfer_b}
}
\caption{The red curves show the results of the training process 
with weight transfer. The blue curves show the results of
the training process with random initialization. 
The left figure shows how the natural logarithm of the
error changes during training.
The right figure shows how the natural logarithm of $||\Delta W||^2_2$ changes 
during training, where $\Delta W$ is the change in $W$ after 100 training steps,
$W$ is the weight matrix.\label{fig:transfer}}
\end{figure}

The network consists of a stack of 3 blocks with m=10, that is, six fully-connected 
layers and three residual connections and a final linear transformation
layer to obtain $u$.  We show how the error and the weights in the layers 
change during the training period in Figure \ref{fig:transfer}. 

%{\bf 3 layers?}

%{\bf I have explained the details about the network architecture.}

%When we start from using truncated normal initializer with $\mu=0.0$ and $\sigma=0.1$ to initialize the weights  and train the network with learning rate=0.01, decay to half every 10,000 steps.

%{\bf I don't understand the description above}

%{\bf The description is not necessary (too many details). I just deleted it.}

We also transfer the weights from the problem:
\begin{equation}
\begin{aligned}
-&\Delta u(x)=0,\quad &x\in \Omega\\
&u(x)=r^{\frac{1}{2}}\sin{\frac{\theta}{2}} \quad &x\in \partial \Omega
\end{aligned}
\end{equation}
where $\Omega=(-1,1)\times (-1,1) \backslash [0,1)\times \{0\}$.

The error and the weights during the training period are also shown in Figure \ref{fig:transfer}.
We see that transferring weights speeds up the training process
considerably during the initial stage of the training.
This suggests that transferring weights is a particularly
effective procedure if the accuracy requirement is not very strigent.

%{\bf where is $W$ defined?}

\subsection{Eigenvalue problems}

Consider the following  problem:
\begin{equation}
\begin{aligned}
-&\Delta u+v \cdot u=\lambda u, \quad x \in \Omega\\
&u|_{\partial \Omega}=0
\end{aligned}
\end{equation}
Problems of this kind occur often in quantum mechanics where
$v$ is the potential function.

There is a well-known variational principle for the smallest eigenvalue:
\begin{equation}
\begin{aligned}
\min \quad &\frac{\int_{\Omega} |\nabla u|^2 dx + \int_{\Omega} vu^2dx}{\int_{\Omega} u^2 dx}\\
s.t. \quad &u|_{\partial \Omega}=0
\end{aligned}
\end{equation}
The functional we minimize here is called the Rayleigh quotient.

To avoid getting the trivial optimizer $u = 0$, instead of using the
functional
$$L_0(x)=\frac{\int_{\Omega} |\nabla u|^2 dx 
+ \int_{\Omega} vu^2dx}{\int_{\Omega} u^2 dx} 
+ \beta \int_{\partial \Omega} u(x)^2 dx
$$ 
we use
\begin{equation}
\begin{aligned}
\min &\quad \frac{\int_{\Omega} |\nabla u|^2 dx + \int_{\Omega} vu^2dx}{\int_{\Omega} u^2 dx}\\
s.t. &\quad {\int_{\Omega} |\nabla u|^2 dx}=1\\
&\quad u|_{\partial \Omega}=0
\end{aligned}
\end{equation}
In practice, we use
\begin{equation}
L(u(x;\theta))=\frac{\int_{\Omega} |\nabla u|^2 dx 
+ \int_{\Omega} vu^2 dx}{\int_{\Omega} u^2 dx} 
+\beta \int_{\partial \Omega}u(x)^2 dx 
+ \gamma \left(\int_\Omega u^2 dx-1\right)^2
\end{equation}
One might suggest that with the last penalty term, the denominator in the
Rayleigh quotient is no longer necessary.  It turns out that we found
in practice that this term still helps in two ways: (1) In the presence of
this denominator, there is no need to choose a large value of $\gamma$.
For the harmonic oscillator in $d=5$, we choose
$\beta =2000$,  $\gamma$ to be 100 and this seems to be
large enough.  (2) This term helps
to speed up the training process.

%{\bf if we include the last term, maybe we can omit the denominator in the first term? this will make the functional simpler}

%{\bf if we keep the denominator in the first term, we do not need to strictly limit $\int u^2 dx=1$, therefore we need not to set a large $\gamma$. Furthermore, the denominator in the first term helps the model converge fast in practice.}

\begin{figure}[!h]
\centering
\vspace{2pt}
%\subfigure[Solution of Deep Ritz method, $811$ parameters]{
\includegraphics[width=0.3\textwidth,height=0.5\textwidth]{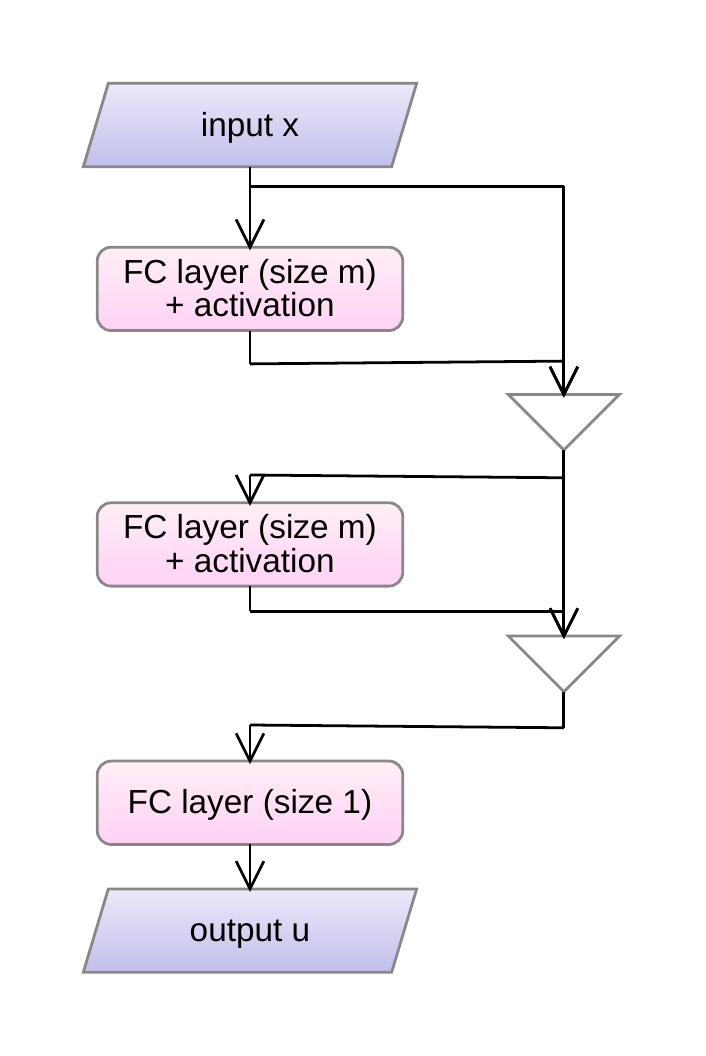}
%\label{fig:net_structure_subfig}
%}
\caption{Network structure used for the eigenvalue problem.
There are skip connections between every pairwise layers. 
The triangles denote concatenation operations.\label{fig:net_structure_2}}
\end{figure}

To solve this problem, we build a deep neural network much like the Densenet \cite{densenet}. 
There are skip connections between every pairwise layers, which help gradients flow through the whole network. The network structure is shown in Figure \ref{fig:net_structure_2}.

%{\bf I don't understand this description.}

%{\bf I have rewritten it and added a figure.}

\begin{equation}
y_i=\phi(W_{i-1}x_{i-1}+b_{i-1})
\end{equation}
\begin{equation}
x_i=[x_{i-1};y_i]
\end{equation}
We use an activation function $\phi(x)=max(0,x)^2$.
If we use the same activation function as before, we found that the gradients
can become quite large and we may face the gradient explosion problem.

%{\bf why not using cubic as before?}

The remaining components of the algorithm are very much the same as before.

\vspace{.1in}
\noindent
{\bf Example 1: Infinite potential well}

Consider the potential function
\begin{equation}
v(x)=\left\{
\begin{aligned}
0, &  \quad x \in [0,1]^d \\
\infty , & \quad x \notin [0,1]^d
\end{aligned}
\right.
\end{equation}
The problem is then equivalent to solving:
\begin{equation}
\begin{aligned}
-&\Delta u=Eu,\quad &x \in [0,1]^d\\
&u(x)=0, \qquad &x \in \partial [0,1]^d
\end{aligned}
\end{equation}
The smallest eigenvalue is $\lambda_0=d\pi ^2$.

The results of the Deep Ritz method in different 
dimensions are shown in Table \ref{tab:ipw}.

\begin{table}[!h]
\caption{Error of deep Ritz method}\label{tab:ipw}
\begin{center}
\begin{tabular}{c|c|c|c}
\hline
Dimension $d$ & Exact $\lambda_0$ & Approximate & Error \\ \hline
1 & 9.87 & 9.85  &  0.20\% \\ \hline
5 & 49.35 & 49.29 &  0.11\% \\ \hline
10 & 98.70 & 92.35 & 6.43\% \\ \hline
\end{tabular}
\end{center}
\end{table}

\vspace{.1in}
\noindent
{\bf Example 2: The harmonic oscillator}

The potential function in $\mathbb{R}^d$ is $v(x)=|x|^2$. 
For simplicity, we truncate the computational domain from
$\mathbb{R}^d$ to 
$[-3, 3]^d$.
Obviously, there are better strategies, but we leave improvements
to later work.

The results in different dimensions are shown in Table \ref{tab:ho}.

\begin{table}[!h]
\caption{Error of deep Ritz method}\label{tab:ho}
\begin{center}
\begin{tabular}{c|c|c|c}
\hline
Dimension $d$ & Exact $\lambda_0$ & Approximate & Error \\ \hline
1 & 1 & 1.0016  & 0.16\%  \\ \hline
5 & 5 &  5.0814 & 1.6\% \\ \hline
10 & 10 & 11.26  & 12.6\% \\ \hline
\end{tabular}
\end{center}
\end{table}
The results deteriorate substantially as the dimension is increased.
We believe that there is still a lot of room for improving the results.
We will leave this to future work.

%{\bf why not compute the per centage for the last item?}

%{\bf I have corrected it.}

\section{Discussion}

We proposed a variational method based on representing the trial
functions by deep neural networks.
Our limited experience with this method suggests that it has the following
advantages:
\begin{enumerate}
\item It is naturally adaptive.
\item It is less sensitive to the dimensionality of the problem
and has the potential to work in rather high dimensions.
\item The method is reasonably simple and fits well with the stochastic
gradient descent framework commonly used in deep learning.
\end{enumerate}
We also see a number of disadvantages that need to be addressed
in future work:
\begin{enumerate}
\item  The variational problem that we obtain at the end is not
convex even when the initial problem is. 
 The issue of local minima and saddle points is non-trivial.
\item At the present time, there is no consistent conclusion about the
convergence rate.
\item The treatment of the essential boundary condition 
is not as simple as for the traditional methods.
\end{enumerate}

In addition, there are still interesting issues regarding the choice of
the network structure, the activation function and the minimization algorithm.
The present paper is far from being the last word on the subject. 

{\bf Acknowledgement:}
We are grateful to Professor Ruo Li and Dr.
Zhanxing Zhu for very helpful discussions.
The work of E and Yu is supported in part by 
%the Major Program of NNSFC under grant 91130005, 
%ONR N00014-13-1-0338, DOE DE-SC0009248.
the National Key Basic Research Program of China 2015CB856000, Major 
Program of NNSFC under grant 91130005,
DOE grant DE-SC0009248, and ONR grant N00014-13-1-0338.

%References {\bf please help to finish the details}:

%{\bf I have finished the details and added some new cited papers.}


\begin{thebibliography}{12}

\bibitem{book}
I. Goodfellow, Y. Bengio and A. Courville, {\em Deep Learning}. MIT Press, 2016.

\bibitem{E}
W. E, ``A proposal for machine learning via dynamical systems'',
{\em Communications in Mathematics and Statistics},
March 2017, Volume 5, Issue 1, pp 1-11.

\bibitem{Han-Jentzen-E}
J. Q. Han,  A. Jentzen and W. E, 
``Overcoming the curse of dimensionality:
Solving high-dimensional partial differential equations using deep learning'',
submitted,
arXiv:1707.02568.

\bibitem{E-Han-Jentzen}
W. E, J. Q.  Han and A. Jentzen,
``Deep learning-based numerical methods for high-dimensional
parabolic partial differential equations and backward stochastic
differential equations'', submitted,
arXiv:1706.04702.

\bibitem{Beck-E-Jentzen}
C. Beck, W. E and Arnulf Jentzen,
``Machine learning approximation algorithms
for high-dimensional fully nonlinear partial differential equations and 
second-order backward stochastic differential equations'',
submitted.  arXiv:1709.05963.

\bibitem{Han-Zhang-Car-E}
J. Q. Han, L. Zhang, R. Car and W. E,
``Deep Potential: A general and ``first-principle'' 
representation of the potential energy'', submitted,
arXiv:1707.01478.

\bibitem{Zhang-Han-Wang-Car-E}
L. Zhang, J.Q. Han, H. Wang, R. Car and W.E,
``Deep Potential Molecular Dynamics:
A scalable model with the accuracy of quantum mechanics'',
submitted,
arXiv:1707.09571.

\bibitem{Evans}
L. C. Evans, {\em Partial Differential Equations, 2nd ed}.  American Mathematical Society, 2010.

\bibitem{resnet}
K. M. He, X. Y. Zhang, S. Q. Ren, J. Sun, ``Deep residual learning for image recognition", {\em 2016 IEEE Conference on Computer Vision and Pattern Recognition (CVPR)}, vol. 00, no. , pp. 770-778, 2016, doi:10.1109/CVPR.2016.90 

\bibitem{adam}
D. P. Kingma, and J. Ba. ``Adam: A method for stochastic optimization." {\em arXiv preprint arXiv:1412.6980}, 2014.

\bibitem{Strang-Fix}
%G.  Strang and G.  Fix, {\em The finite element method}, ...
G. Strang and G. Fix, {\em An Analysis of the Finite Element Method}. Prentice-Hall, 1973.

\bibitem{densenet}
G. Huang, Z. Liu, K. Q. Weinberger, V. D. M. Laurens, ``Densely connected convolutional networks.", {\em arXiv preprint arXiv:1608.06993}, 2016.
\end{thebibliography}
\end{document}